\documentclass[11pt]{article}

\usepackage[final]{acl} 
\usepackage{graphicx}

\usepackage{tabularx, booktabs}
\usepackage{enumitem}
\usepackage{array}

\usepackage{times}
\usepackage{latexsym}

\usepackage[T1]{fontenc}

\usepackage[utf8]{inputenc}

\usepackage{microtype}

\usepackage{cleveref}
\crefformat{section}{\S#2#1#3} 
\crefformat{subsection}{\S#2#1#3}
\crefformat{subsubsection}{\S#2#1#3}

\newcommand{\unesco}{{\sc{unesco}}}

\title{Modeling the Sacred: 
Considerations when Using Religious Texts \\ in Natural Language Processing}

\author{Ben Hutchinson \\
  Google Research, Australia \\
  \texttt{\small benhutch@google.com}
  }

\begin{document}
\maketitle
\begin{abstract}
This position paper concerns the use of religious texts in Natural Language Processing (NLP), which is of special interest to the Ethics of NLP. Religious texts are expressions of culturally important values, and machine learned models have a  propensity to reproduce cultural values encoded in their training data. Furthermore, translations of religious texts are frequently used by NLP researchers when language data is scarce. This repurposes the translations from their original uses and motivations, which often involve attracting new followers. This paper argues that NLP's use of such texts raises considerations that go beyond model biases, including data provenance, cultural contexts,  and their use in proselytism. We argue for more consideration of researcher positionality, and of the perspectives of marginalized linguistic and religious communities.
\end{abstract}

\section{Introduction}
\label{sec:intro}

\begin{quote}
\emph{``a particular string of speech may be viewed as data by a researcher but as sacred incantation by language users''} \\
--- \citet{holton2022indigenous}
\end{quote}


The Association for Computational Linguistics (ACL) is a secular institution. Its constitution, resolutions and policies make no mention of religion other than forbidding harassment on the basis of religion.\footnote{\url{www.aclweb.org}, accessed September 2023} Nevertheless
the Christian Bible and the Islamic Quran\footnote{This paper follows several style guides in using ``Quran'', although mentions of the alternate Latinization ``Koran'' are also considered in the corpus studies we report on.} are often used in the scientific and professional activities of ACL, as measured by papers published in the ACL Anthology (see Section~\ref{sec:texts in NLP}). 
Some of the reasons that NLP researchers use the Bible are aptly expressed by \citet{resnik1999bible}.
The Bible is the world's most translated book, with translations in over 2,000 languages, and often multiple translations per language. Furthermore, great care is taken with the translations, so from an NLP perspective data quality is high. It is often easily available in electronic form, and is in the public domain, hence free to use. It has a standard structure which allows parallel alignment verse-by-verse.
For these reasons, as recently as 2006 it was said to be ``perhaps surprising that the Bible has not
been more widely used as a multilingual corpus
by the computational linguistics and information
retrieval community'' \cite{chew2006evaluation}.

Despite the use of religious texts in NLP, 
the ethical considerations of such use have not previously received much attention. Religious texts have generally been treated as just data, overlooking their sacred dimensions, their cultural significance, and their histories, which are sometimes entwined with colonial projects. 
This position paper contends that responsible secularism demands engaging with the ethical considerations of the use of these texts. 
\begin{table*}[t]
    \centering
    {\small
\begin{tabular}{llrc}
    \toprule
Religion        & Sacred texts                             & Est.\ 2020 population  & Proselytizing\\
\midrule
Christianity    &Bible, New Testament, Old Testament    & 2,382,750,000 & Yes\\ 
Islam           &Quran (alt.\ spellings include Koran)    & 1,907,110,000 & Yes\\
Hinduism        &Vedas, Upanishads, Puranas             & 1,161,440,000 & No\\
Buddhism        &Tripitaka, Mahayana Sutras, Tibetan Book of the Dead & 506,990,000 & Yes \\
Traditional Chinese Religion	        &Zhuangzi, Tao-te Ching, Daozang        &310,000,000    & No\\	
Judaism         &Talmod, Torah, Tanakh, Old Testament   & 14,660,000 & No \\ 
\bottomrule
\end{tabular}
}
    \caption{Some major world religions and their texts. Population estimates are from the US-based Pew Research Center (\url{www.pewresearch.org}), which conducts demographic and other research.}
    \label{tab:world religions}
\end{table*}
%
We take the position that it is important to contextualize religious texts within not just their religious but also their broader cultural and historical contexts.
We argue that it is important to acknowledge the motivations, goals, and positionalities of researchers using religious texts, particularly as it is difficult to have intimate first-hand knowledge of the moral and cultural concerns of a religion beyond one's own (if any).

Section~\ref{sec:background} provides relevant background material, including discussing the relationship between academic linguistics and missionary linguistics. In Section~\ref{sec:texts in NLP}, we present an empirical study of NLP papers using sacred texts. Section~\ref{sec:considerations} discusses various ethical considerations when using sacred texts in NLP. In doing so, we consider a range of approaches to the topic, including ethical theories, Indigenous perspectives,\footnote{Following style guides such as \cite{younging2018elements}, this paper capitalizes the first letter of ``Indigenous''.}
human rights, and the AI principles commonly espoused by institutions.
Our goal in doing so is not to evaluate past NLP projects that use religious texts, but rather to encourage more reflecting in and on future work.
Based on these considerations, we then make some recommendations for the NLP community in Section~\ref{sec:reccommendations}.

\section{Background}
\label{sec:background}

\subsection{Religion}
Precisely defining what constitutes a religion might be notoriously difficult \cite[see, e.g., ][]{spiro2013religion, neville2018defining}, and lies beyond the scope of this paper.  
Typical properties of religions center around giving meaning to existence, and include moral values, spiritual beliefs,  theistic beliefs, rituals, stories and mythologies, kinship systems and marriage practices,
artistic practices, significant locations, and in some cases a language which plays a special role. These are all closely related to questions concerning values.

Like languages, religions may exhibit regional variation and incorporate local practices, so thinking of them as discrete entities may be somewhat misleading. Some widely cited estimates put the number of worldwide religions at several thousand, although these claims are disputed. What seems more certain is that the imminent extinction of thousands of Indigenous languages will accompany an ``impending loss of so many religions and worldviews'' \cite[p.\ 153]{harrison2007languages}. 
Acknowledging these challenges, we nevertheless provide a summary of some of the world's most populous religions in Table~\ref{tab:world religions}.
Surveying or defining each of these religions is beyond the scope of this paper. Also not included here are the various sects and branches within each religion, nor texts which might be important only to specific branches.

One important distinction is that between proselytizing and non-proselytizing religions. The former attempt to convert new populations, whereas the latter do not. The former are more intricately related to historical practices of colonialism---especially in Africa, the Americas, Asia, Australia, and the Pacific---and hence also to neocolonial legacies. Some religions hold that a certain language is privileged for communicating sacred texts to the faithful, while on the other hand Protestant Christianity exemplifies a commitment to communicating in vernacular languages.
(Article XXIV of the Articles of Religion of the Anglican Church calls for ``such a Tongue as the people understandeth''.) 
Although originally in Ancient Greek, Christian texts have been widely distributed in many European languages since the Protestant Reformation, and the global reach of Christianity is associated with European colonial practices. 


\subsection{The academy and Bible translation}

The September 2009 issue of the journal \emph{Language} has a special feature of five articles by anthropologists and linguists concerning the relationships between the US-based Bible translation organization SIL International (SIL) and academic linguistics. In this issue, \citet{dobrin2009} explore how academic linguists have at times become reliant on, and benefitted from, the technological infrastructures of SIL, in part because creating and maintaining these infrastructures have not been valued by the academy. This ``partnership of convenience'' causes tensions between differing objectives, and raises questions about what kind of relationships secular research institutions ought to have with organizations with very different agendas. These practices presage similar ways in which some areas of NLP research have become reliant on Bible translations. 

Many linguists and NLP practitioners working on Indigenous languages see their research as addressing issues of human rights and cultural extinction. However, as Dobrin and Good point out, languages which are most endangered are least likely to receive SIL's attention. \citet{handman2009language} draws attention to how SIL ideology separates linguistic identity from religious identity, differing from \unesco's position that sustaining endangered languages entails sustaining cultural worldviews, knowledge systems, and identity practices. \citet{epps2009syntax} argue that evangelical success entails the displacement or transformation of traditional beliefs, often leading to social upheaval, and argues that the academy has a moral interest in supporting local self-determination which is at odds with evangelical agendas.

\section{The Use of Sacred Texts in NLP}
\label{sec:texts in NLP}

This section presents empirical data on the use of religious texts in the field of NLP.

\subsection{Sacred texts in the ACL Anthology}
\label{sec:anthology}

In this section we consider four research questions: i) \emph{which} sacred texts are used in NLP?, ii) how does NLP \emph{characterize} those sacred texts?, iii) for what purposes does NLP \emph{use} sacred texts?, and iv) are there any \emph{trends} over time?
To explore these, we use the Anthology of the Association of Computational Linguistics\footnote{\url{www.aclanthology.org}, accessed August 2023.} (henceforth, the ACL Anthology) \cite{bird2008acl}, which we consider to represent the research publications of the NLP community.

Regarding \emph{which} sacred texts are used in NLP, the number of ACL Anthology entries for sacred texts are shown in Table~\ref{tab:religious texts}.\footnote{Non-deterministic result counts seem to be an artefact of the ACL Anthology's use of Google's Programmable Search Engine \cite{bollmann2023two}.} Thousands of papers in the ACL Anthology seem to use religious texts. There is a strong bias towards the texts of the Judaism, Christianity, and Islam. 

\begin{table}[t]
    \centering
    {\small
    \begin{tabular}{lrlr}
    Search term/phrase & Results [min, max] \\ 
    \midrule
bible	& [1920, 3890]] \\
quran	& [291, 547]  \\ 
new testament 	& [294, 294] \\
koran	& [131, 248] \\
old testament 	& [73, 206] \\
torah	& [25, 153] \\
talmud 	& [21, 22] \\
vedas 	& [22, 51]\\
tripitaka	& [7, 7]  \\
upanishads	& [6, 6] \\
mahayana sutras	& [4, 4] \\
tanakh 	& [3, 4] \\
zhuangzi	& [3, 3] \\
puranas	& [3, 3] \\ 
tibetan book of the dead	& [0, 0] \\
tao-te ching	& [0, 0] \\
daozang & [0, 0]\\
\bottomrule
    \end{tabular}
    }
    \caption{Number of search results for religious texts in the ACL Anthology on August 10, 2023. Since search result counts are non-deterministic, we report the min and max of 10 searches for each term. 
    }
    \label{tab:religious texts}
\end{table}

To understand how sacred texts are \emph{characterized} in NLP, we analysed ACL Anthology papers containing the phrases ``the Bible is'' or ``the Quran is''. Thematic analysis of the results reveal that descriptions of the Bible mostly emphasize its availability, convenience, size, and multilingual character (see Appendix A). The religious nature of the Bible is only mentioned  infrequently, in stark contrast with how Christian websites describe the Bible. These findings agree with the claim that NLP has a dominant ideology of ``language as data'' \cite{bird2024centering}.  One paper describes the Bible as ``one of the most familiar documents''. NLP papers are more likely to mention religious aspects of the Quran. No ACL anthology papers mention ``the Koran is'', nor related appositions ``Koran, a[n]''; presumably the papers mentioning `Koran' felt it needed no explanation.

To understand how NLP \emph{uses} sacred texts, we examined the first 100 results (sorted by relevance) from the ACL Anthology for each of the terms `bible' and `koran'. We manually looked at each of the 200 search results in order to verify its relevance to our study, and  omitted 48 search results for not being relevant (see Appendix~B). We manually coded the 88 papers mentioning the Bible for their application domain.  The most common application domain was machine translation (23\%), while many (18\%) introduced a new corpus or lexical resource. Three papers were concerned specifically with Bible translation, and four with literary analysis of the Bible. A variety of other application domains were represented, including various morpho-syntactic analysis tasks, and language modeling or pretraining (see Appendix C).  48\% of the 88 papers concerned one or more Indigenous languages.\footnote{These categorizations were done by the author, taking into account historical and social context, however an ideal approach might engage with language communities to understand whether they consider themselves to be Indigenous.} The 64 ACL papers mentioning `koran' were less varied and largely (73\%) were concerned with machine translation, typically using verses from the Quran in evaluation datasets (see Appendix D). 

\begin{figure}[t]
    \centering
    \includegraphics[width=3.5cm]{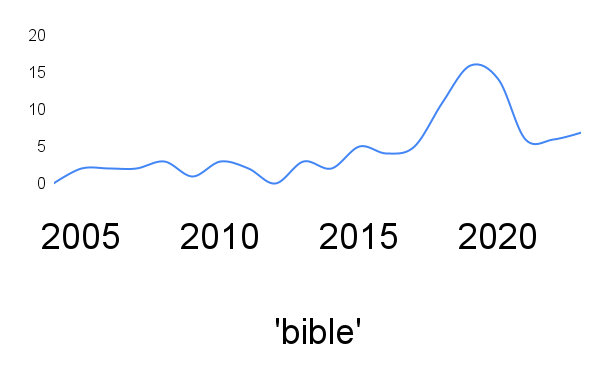}
    \includegraphics[width=3.5cm]{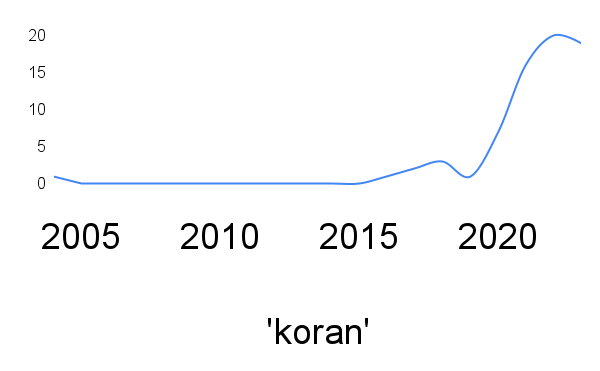}
    \caption{Trend sparklines for counts of papers in the ACL Anthology mentioning `bible' or `koran' (see 
    \S\ref{sec:texts in NLP}).}
    \label{fig:bible by time}
\end{figure}

To get a sense of \emph{trends} in the use of sacred texts by the NLP community, we examined the year of publication of the 150 of the  152 papers described above which were published between Jan 2004 and July 2023 (just 2 were published prior to 2004). Of these, we see increasing use over time of mentions of the Bible, and especially of mentions of `Koran' (see Figure~\ref{fig:bible by time})\footnote{Since our 2023 sample was limited to Jan-July, paper counts for 2023 are multiplied by $12/7$ when creating sparklines, to make them comparable with previous years.}, including over 60\% of papers being published between 2019 and 2023 alone.
Of course, the number of papers published in the ACL anthology has also been increasing, for example ACL 2023 accepted about 12 times as many papers as ACL 2004 did. We are not necessarily seeing an increase in terms of the fraction of NLP papers using the Bible, and thus there may only be weak evidence of trends \emph{within} NLP research. We do however hypothesize that the absolute counts of papers using religious texts may be a better proxy for the \emph{externalities} (both positive and negative) of NLP research on linguistic and religious communities, and thus relevant to this paper's topic of ethical considerations.  
There also seems to be a trend towards papers using sacred texts handling very large numbers of languages, with ten papers since 2014 in our sample handling over 500 languages.

\subsection{Four case studies}
\label{sec:case studies}

To complement the broad analysis of NLP research above, we also report here on some recent noteworthy cases which have attracted acclaim and attention.
These illustrate in more detail ways in which the NLP community is encountering and using religious texts.

Our first example is a paper which aims to improve Speech Recognition and Speech Synthesis for over a thousand languages  \cite{pratap2023scaling}. It was uploaded to arXiv, a popular online archive for computer science papers, in May 2023.
The researchers train their model using translations of the New Testament, as well as audio of readings of those translations, obtained from Faith Comes by Hearing (\url{faithcomesbyhearing.com}), \url{goto.bible} and \url{bible.com}. They also use spoken recordings in many languages, without paired texts, of Bible stories, evangelistic messages, scripture readings, and songs, obtained from Global Recordings Network (\url{globalrecordings.net}), whose mission is to communicate ``the Good News of Jesus Christ'' via a strategy of recording, distribution, and promotion.

Our second example aims to scale language models to 500 languages \cite{imanigooghari2023glot500}, and  was awarded the ACL Area Chair Award for best Multilingualism and Cross-Lingual NLP paper in July 2023. 
The researchers ``crawl or download'' data from 150 sources, including religious texts and observe a ``higher proportion of religious data'' compared to previous comparable work. Parallel verses from Bible translations are used for model training and testing, and performance is reported for Sentence Retrieval from the Bible. 

Our third example concerns the use of JW300 \cite{agic2019jw300}, a dataset of around 100k sentences in each of 300 languages crawled from \url{jw.org}, a website run by the US-based Jehovah's Witnesses, a Christian denomination. A majority of the texts come from the Jehovah's Witnesses' magazines \emph{Awake!} and \emph{Watchtower}. Released as a corpus in 2019, JW300 has been cited over 180 times as of March 2024. The African grassroots open-source NLP project Masakhane (\url{masakhane.io}) had been using JW300 to train Machine Translation models, until receiving legal advice in 2023 that this was breaching copyright. A subsequent request by Masakhane to the Jehovah's Witnesses for permission to use the data was declined.\footnote{See \url{https://infojustice.org/archives/45258}.}


Our final recent example concerns the release of {\sc madlad-400}, a new text dataset containing 3T tokens in 419 languages \cite{kudugunta2023madlad}. It uses 2022 snapshots of the CommonCrawl web crawl (\url{commoncrawl.org}) and the paper was uploaded to arXiv in September 2023. Auditing of a preliminary version of the dataset, spanning 498 languages, revealed that  for 141 languages there were ``significant amounts'' of Bible data. Significant amounts of Jehovah's Witnesses data was also found for 37 languages, and of Church of Jesus Christ of the Latter Day Saints (LDS) data for 2 languages. (No Quran data was reported to be found in significant amounts.)

\section{Considerations around Sacred Texts}
\label{sec:considerations}

Having demonstrated above that religious texts have been used in thousands of NLP papers, we now discuss some of the ethical considerations. This topic has not received much prior attention, although it is touched upon by \citet[sec. 3]{mager2023ethical} and \citet[sec.\ 8]{pratap2023scaling}.
Our goal is not to critique the works identified in the previous section, but rather to provide a toolbox for assisting critical thinking in the future.
In this section we lay out our positions that relevant considerations include not just technological outcomes, but also cultural standpoints, cultural inclusion, and relationships between global and marginalized cultures. With this in mind, we now consider the use of sacred texts in NLP from a range of lenses. 

\subsection{Ethical theories}
\label{sec:ethical theories}

Ethical theories are attempts at explaining why actions might be called right or wrong. In this section we consider the use of sacred texts through just two ethical theories, however our broader point is that if analyses of NLP projects focus narrowly on using a single theory then they risk minimizing concerns which other theories bring more focus to.   

The term \textbf{consequentialism} refers to a family of normative theories which emphasize the importance of considering the consequences of actions, including analysis of risks and benefits. One example of a consequentialist theory is \textbf{utilitarianism}, which holds that the best action is one that maximizes wellbeing and minimizes suffering.
One challenge with applying these theories in an NLP research context is that in practice research activities are often far removed from applications, and any eventual path between the two can be unknowable or uncertain. Even if the impacts on users of NLP applications can be known, challenges arise in calculating aggregate benefits across disparate stakeholders with different objectives. 
Perhaps the easiest consequences to reason about are the impacts of NLP research and applications on the lives of the researchers and developers themselves, since research papers and software bring rewards in the forms of measurable kudos, citations, career progression, etc.
How can such benefits be weighed against uncertain risks of causing serious offense to religious communities? 
If translations of religious texts are purchased from a proselytizing organization, what are the downstream impacts of the organization's re-investing those payments into their projects? 

\begin{table*}[t]
    \centering
    \begin{tabularx}{\textwidth}{lX}
    \toprule
         AI Principle& Example possible considerations  \\
         \midrule
         {Safety} &Could the errors made by an MT model trained on sacred texts perpetuate harmful misinformation and hate speech about religiously significant topics? \\
         {Privacy} &In speech-to-speech translation, could spoken recordings of religious texts (cf.\ Section~\ref{sec:case studies}) enable speakers to be identified and their religious beliefs compromised? \\
         {Bias and fairness} &In what specific ways are religious texts not representative of other domains \cite[cf.][]{mayhew2017cheap, adelani2021effect, evans2004searching}? Do these biases cause MT models to have unfair impacts?  What are the specific limitations of MT eval sets which are largely constituted by religious texts?  \\
         {Accountability} &What mechanisms exist for non-English speaking religious communities to report and escalate offensive translations resulting from the use or religious texts? Who is responsible for responding to such reports and mitigating the harms?\\
         {Transparency} &Do we have accessible evidence that religious communities were consulted when building corpora of parallel religious texts to use for MT? \\ 
         \bottomrule
    \end{tabularx}
    \caption{Non-exhaustive list of possible AI Principles considerations related to using religious texts for MT. }
    \label{fig:mt principles}
\end{table*}

The term \textbf{deontology} refers to a normative theory which posits that there are rules or principles which determine the rightness of wrongness of actions, rather than the consequences. Within an NLP context, this might lead to a focus on the upstream actions such as sourcing of data, rather than downstream actions such as usage of NLP applications. 
Within the context of religious texts and their translations, this might lead to questions such as: Were the appropriate people involved in the creation of a dataset of sacred texts? Did they have the right roles and relationships from within the perspective of followers of the religion? Do translators of sacred texts have the right specialist linguistic, religious, and and cultural knowledge?
Were the translators paid fairly, both individually and collectively?
Were consent practices around voice recordings followed?
Generally, was the dataset collected in a manner aligned with best research practices, e.g., as operationalized by research ethics boards? 
\citet{prabhumoye2020case} discuss the importance of informed consent for deontological approaches to NLP ethics, and community-level consent might be an appropriate lens for thinking about communities of religious practice.

\subsection{AI principles}
\label{sec:ai principles}


We use the term \textbf{AI principles} here to refer broadly to the hundreds of sets of principles for responsible and ethical AI that have been released by companies and governments in recent years. Unlike the ethical   theories described above, these are focused on AI technologies.  Although all unique in their own way, the various sets of principles also have many facets in common \cite{floridi2022unified}, such as safety, privacy, bias, fairness, accountability, and transparency.
We expect the AI Principles-related concerns to be highly dependent on NLP system functionality and thus on the application domain.

Given the frequent use of sacred texts for training and evaluating MT systems, we provide a sketch in Table~\ref{fig:mt principles} of how AI Principles might raise different flavors of considerations.\footnote{\citet{mager2023ethical} also provide a useful discussion of Ethics and Machine Translation.}  
Relevant to the potential for MT bias, \citet[][p.\ 4]{evans2004searching}  observe that parallel Bibles will provide signal on how to translate ``arise!'' and ``Cain fought with Abel'', but will not help with translating culture-specific concepts of the world's diverse languages. Similarly, MT evaluation will have limitations whenever eval sets are largely constituted  by parallel religious texts.

\subsection{Cultural considerations}
\label{sec:culture}

Authors of NLP publications do not represent global diversity \cite{rungta2022geographic}, and hence special consideration needs to be given to cultural standpoints.
A useful starting premise might be that ethical consideration of ACL papers  should not be biased either against nor towards any religions. This might seem to suggest that \textbf{Rawls' Veil of Ignorance} could provide a guide \cite{john1971theory}. That is,  one's consideration of the use of sacred texts within NLP should proceed as if we are each in ignorance of which is our own religion (if any). For example, how would I feel about NLP's reliance on texts from major religions if my own culture and religion might be ones which are marginalized and endangered?

Although the Veil of Ignorance might seem attractive at face value, we take the position that our ability in practice to truly avoid being informed by our own cultural backgrounds and affiliations is extremely questionable.
We must instead accept that we have cultural standpoints.  
The \textbf{emic/etic distinction} originated in linguistics in the 1950s for describing different standpoints for language research \cite{mostowlansky2020emic}. \emph{Emic} is commonly used to describe research on a culture from the perspective of people of that culture. This contrasts with \emph{etic} research, which takes an outsider's perspective. When NLP handles religious texts, we can distinguish research problems and applications which are within the researcher's own religious context, from those applications which impact those having other religious beliefs (e.g., translation for the purpose of proselytizing). 

\subsubsection{On relativism and the ACL community}

One possible objection to the argument that greater consideration of NLP's use of sacred texts is needed
is based on an argument of cultural relativism. Such an argument would contend that by using sacred texts for an extended period, the ACL community has demonstrated that such practices are judged as acceptable by the norms of the ACL community.

We counter that such an argument would be stronger if the ACL community both had a stronger history of reflexive practices, and was more culturally diverse. Compared to many other disciplines, we find that ACL's interest in ethics to be relatively recent (see Appendix E).
The ACL adopts the ACM Code of Ethics,\footnote{\url{https://www.aclweb.org/portal/content/acl-code-ethics}} a general code for computing professionals which makes no mention of working with cultural data such as language.   
Unlike some other disciplines, positionality statements are rare in ACL papers, with ``researcher positionality'' and ``author positionality'' each having only a single (and recent) result in the ACL Anthology as of September 2023.  We also find that the ACL community does not represent the diversity of the world's local languages and religions; in fact disparities in sources of ACL publications might be increasing \cite{rungta2022geographic}.

\subsection{Power and marginalization}
\label{sec:marginalized}

The more different that a culture is from the global cultures which dominate NLP research, the less we should expect the NLP community to understand  local concerns.
Consideration of marginalized local cultures might bring our focus to questions of how NLP technologies serve to re-arrange power. Which actions are encouraged or discouraged by NLP technologies, and by who? 
Blodgett has called for NLP researchers to focus their ethical considerations on power relations between technologists and communities \cite{blodgett2020language}. As discussed in Section~\ref{sec:texts in NLP}, many NLP papers using the Bible use translations of its texts into Indigenous languages. In this section we give special attention in this section to concerns of   local Indigenous communities. Their voices are among the least likely to be represented in ACL's prestigious conferences  and journals. And their relationships with global projects are likely to involve large power disparities such that ``the asymmetry of power is the cause of domination'' \cite{mager2023ethical}.



\subsubsection{Indigenous perspectives}

Indigenous concerns regarding \textbf{data sovereignty} describe the importance of data governance and Indigenous Cultural and Intellectual Property (ICIP), but also a broader set of concerns regarding practices for projects involving Indigenous data \cite[e.g., ][]{carroll2020care,national2018ethical,taiuru2021kaitiakitanga, janke2019true, smith2021decolonizing}. Surveying  the plurality of perspectives is beyond the scope of this paper, however they share common themes around respect, cooperative relationships which include consultation and negotiation, shared control, and providing benefits to community \cite{mager2023ethical, cooper-etal-2024-things}.

\subsubsection{Human rights}

As tools of colonizing projects, translations of sacred texts into Indigenous languages have been described as a ``well documented example of the non-ethical misuse of translation'' \cite{mager2023ethical}. 
One consideration for NLP is whether using such translations constitutes complicity with, and promotion of, projects which might violate international human rights to maintain Indigenous cultures. 

Kenyan human rights scholar Makua Mutua describes the ``basic contradictions'' between proselytizing religions and Indigenous cultures \cite{mutua2004proselytism}. Observing that religion is woven into every aspect of Indigenous social and cultural life, including dances, storytelling, and marriage practices, Mutua argues that such cultures'  meeting with proselytizing Christian and Islam faiths amounts to cultural genocide. In some cases, this characterization seems valid.
In other cases, it might perhaps be too strong for the complex realities of syncretic responses by Indigenous communities to proselytizing cultures.

Mutua argues that the right to freedom of religious belief cannot be considered to exist in a level playing field in which local cultures can compete with global ones. Rather, the contexts of cultural invasion unfairly privilege global religions, including histories of missionaries making access to education and health services conditional on the ``salvation'' of ``infidels''. This echoes arguments by legal scholars that the power and sovereignty dynamics between source and target cultures constitute important factors between proper and improper proselytism \cite{stahnke1999proselytism}.
The Human Rights Committee has acknowledged that the cultural rights protected under Article 27 depend on the ability of a minority group ``to maintain its culture, language or religion''.\footnote{General Comment No\. 23, UN Doc.\ ICCPR/C/21/Rev.1/Add.5 (1994), para.\ 6.2.} As such, Prof.\ Mutua argues that the (then Draft)  UN Declaration on the Rights of Indigenous People appears to prohibit proselytizing by agents external to the the Indigenous culture in order to create space for Indigenous peoples to maintain their cultures amidst external threats.

\section{Discussion and Recommendations}
\label{sec:reccommendations}

We take the position that the ACL should strive to be aware of risks of harms to religious communities, including dignitive harms, that NLP research may cause or be complicit with.
It is beyond the scope of this paper to weigh possible benefits of NLP technologies developed using religious texts against the ethical considerations of such use, and any such benefits may be context- and application-specific. However, we note that benefits are often asserted for languages for which NLP has few resources, and we advocate that such claims of benefits should be evaluated  paying attention to local language ecologies \cite{bird2022local}, to the roles of technology in supporting minority languages \cite{holton2011}, and to the real technology needs of communities \cite[e.g.,][]{liu2022not}.

We now suggest five specific actions for the NLP community to mitigate the risks of consequential, procedural, and dignitive harms.

\paragraph{1. Discuss ethical considerations.} If one should never speak of religion in polite company, then perhaps ACL forums should be less polite.
We suggest that discussion of ethical considerations in NLP papers using religious data should be much more prevalent than it currently is. We propose that Section~\ref{sec:considerations} provides a good starting point for NLP researchers to think about possible concerns, however, building on HCI best practices we also encourage researchers to talk to members of relevant religious communities in order to understand concerns which might lie outside the researchers' domains of lived experience.   

\paragraph{2. Consider different ethical theories.} It has been argued that NLP research often implicitly adopts a utilitarian lens, in the process minimizing other ethical considerations  \cite[e.g.,][]{hutchinson2022evaluation}. We encourage NLP researchers using sacred texts to contemplate approaching ethical considerations from numerous angles, including not just deontology (discussed in Section~\ref{sec:ethical theories}), but also alternatives such as virtue ethics and ethics of care.  

\paragraph{3. Delve into domain specifics.} We encourage more work on specific AI principle risks of different NLP domains when using religious texts (cf.\ Section~\ref{sec:ai principles}). What are the specific ways in which various religious texts are not representative of other domains? 
What are the model biases that result? Are there specific risks around offensive misinformation on religious topics? Have datasets incorporating sacred texts followed best practices around transparency \cite[e.g.][]{bender2018data,gebru2021datasheets,pushkarna2022data}? 

\paragraph{4. Situate NLP work culturally.} When should an NLP practitioner tread more carefully? We suggest that when handling sacred data from other cultures and religions there will be more risks, since the NLP practitioner will have almost certainly have gaps in their knowledge, and will likely also have differing values \cite{hershcovich2022challenges,prabhakaran2022cultural}. We suggest it is useful to talk of \emph{etic NLP} and \emph{emic NLP} (cf.\ Section~\ref{sec:culture}), according to whether the language technology is for the own linguistic and religious cultures of the NLP practitioner vs.\ those of others. Linguistic positionality statements have been recommended by researchers who are aware of how different priorities and agendas between researchers and language communities can impact projects \cite[e.g.,][]{rolland2023methodological, cormier2018language}.  Similarly, we suggest that {religion positionality statements}, for NLP research working with religious data, can also provide a useful signal for the NLP community concerning agendas. For example, hypothetically, would non-Muslim researchers understand the Muslim community's concerns regarding automated mis-translations of the Quran?

\paragraph{5. Attend to concerns of marginalized cultures.}
Given the NLP community's skew towards the West \cite{rungta2022geographic}, the values of, and possible harms to, local communities of diverse cultures are not known by most NLP researchers. For Indigenous language NLP projects using translations of sacred texts, we echo calls for more consideration of local colonial contexts, to consider community opinions, and for research to prioritize the needs of Indigenous communities   \cite{bird2020decolonising,schwartz2022primum,alvarado2021decolonial}. \citet{mager2023ethical} demonstrate one way in which community opinions can be sought regarding NLP projects, and deeper relationships with communities will provide more insights \cite{bird2024centering, cooper-etal-2024-things}. NLP researchers working with Indigenous languages should become familiar with Indigenous perspectives mentioned in Section~\ref{sec:marginalized}, and with codes of conduct such as that of the Endangered Languages Project.\footnote{\url{https://fpcc.ca/wp-content/uploads/2023/02/CodeOfConduct_Web.pdf}}
With the imminent extinction of many Indigenous languages and religions, we also suggest there may be a role for NLP to play in documenting and maintaining ``first-person accounts of what people once believed in and how they talked to and about their gods'' \cite[][p, 153]{harrison2007languages}.

One pertinent question is how often the papers surveyed in Section~\ref{sec:texts in NLP} adhere to these recommendations. Regarding \textbf{discussing ethical considerations}, \citet[sec.\ 8.3]{pratap2023scaling} do include a section, albeit short, on Ethical Considerations of the use of religious texts, although we found that doing so is rare. \citet{kudugunta2023madlad} and \citet{imanigooghari2023glot500} both have short ethics statements, but neither mentions  religion. Despite translation of the Quran being a ``problematic and controversial issue for Muslims'' \cite{fatani2006}, none of the many papers surveyed in Section~\ref{sec:anthology} that evaluate MT systems discuss ethical considerations.  

Regarding \textbf{ethical theories}, papers using religious texts do not seem to break the general NLP trend of adopting a utilitarian mindset \cite{hutchinson2022evaluation} and focusing ethical discussion on narrow questions such as decontextualized bias measurement \cite{blodgett2020language}. 

Regarding \textbf{domain specifics}, \cite{pratap2023scaling} attempt to understand biases in their data, although the details are scarce, and their comparison is limited to a dataset based on English language Wikipedia.  \citet{kudugunta2023madlad} and \citet{imanigooghari2023glot500} have no discussions of biases, although the latter do include a datasheet \cite{gebru2021datasheets}.

Regarding \textbf{situating work culturally}, positionality statements are extremely rare in the ACL Anthology papers. Some rare examples of NLP papers mentioning author positionality or standpoint are \cite{mcmillan-major-etal-2021-reusable, hutchinson-etal-2022-underspecification,santy-etal-2023-nlpositionality, yoder-etal-2023-weakly, ghosh2023chatgpt}.
Only half of ACL papers even mention the language(s) they study \cite{ducel2022we}, giving a false veneer of independence \cite{bender2019benderrule}. 

Regarding \textbf{attending to concerns of marginalized cultures}, this is rarely done in NLP research. NLP papers that consider ethical topics tend to focus on English, and most papers on languages other than English focus on model accuracy \cite{ruder-etal-2022-square}.    \citet{pratap2023scaling} mention how their bias analysis was constrained by lack of easily available native speakers for most languages, which also precluded the possibility of discovering community concerns beyond bias.

\section{Conclusion}

Thousands of NLP research papers have used religious texts, due to their availability, convenience, and multilinguality.
However these papers rarely acknowledge the religious significance of these texts, nor their cultural contexts, and the ethical considerations of such use have not previously been explored.
This position paper presented the first detailed study of the use of religious texts in NLP research, finding that common scenarios include  machine translation and dataset creation. We have argued that responsible secularism requires the NLP community to engage with concerns about how such NLP activities might impact religious communities---especially the most marginalized ones---or might be complicit with projects which do. We provided a detailed account of some of the considerations, with a focus on ethical theories, AI principles, cultural considerations, and marginalized communities. We suggested that the field of NLP would benefit from  more discussion of the diverse and specific concerns around using religious texts. We proposed that the NLP community  engage more with questions of researcher positionality and cultural standpoints, especially with regards to marginalized cultures.

\section{Researcher Positionality}
I live and work in a secular, multicultural, majority English-speaking, colonized country of the Global North.
I am linguistically competent only in English. My experiences include studies and employment in linguistics and NLP, but I have no background in religious studies. 
I grappled while writing this paper with my lack of first-hand experiential understanding of religion, and thus too with my personal role in arguing for more consideration of global and marginalized religious communities.


\section*{Limitations}

Any position paper is limited by the experiences of the authors. A Researcher Positionality statement is included partly to address this.

We acknowledge that the ACL Anthology used in Section~\ref{sec:texts in NLP} might not be representative of the entire field of NLP, e.g., missing relevant work such as \cite{chandra2022artificial,bashir2023arabic}, and NLP projects in industry might not be well represented in the ACL Anthology. Another limitation is that we exclude publications in languages other than English.  Our reliance on keyword searches, rather than more sophisticated analyses of the ACL corpus, is a limitation. We also acknowledge that the ACL Anthology's non-determinate search result counts are a limitation, seemingly an artefact of the use under the hood of Google's Programmable Search Engine, however reproducing the ACL Anthology's search functionality is beyond the scope of this position paper.

This paper was unable to accommodate lengthy discussion of religious texts in specific NLP domains or specific NLP projects. As the topic of religious texts in NLP has not previously received attention, we took the position that focusing on breadth was most useful. However in Section~\ref{sec:reccommendations} we do suggest that future work focus more on ethical considerations of using religious texts in specific NLP domains. 

This paper does not do justice to the complexities of many of the topics it touches upon. Perhaps chief among these are the histories of Indigenous communities, proselytizing religions and colonial projects. Colonial missionaries saw their own roles in a positive way despite the accompanying cultural destruction \cite{nakata2007disciplining}.
However in recent decades, there have at times been apologies from some church bodies for histories of religious domination which marginalized Indigenous languages and religious practices \cite{bush2015canadian}.
Indigenous communities have at times, facing many colonial pressures including threats of language extinction,  used Bible translations as tools for language maintenance \cite[e.g.,][]{davis2015intersections}. Cultural protocols can also be maintained, for example  spoken recordings of Bible translations in Kanien’kéha are still subject to cultural practices around speakers who have passed away \cite{pine2022requirements}, while on Groote Eylandt, in Northern Australia, the Anindilyakwa people reinterpreted Christianity of the missions in their own ways leading to a ``hybridisation of cultures'' \cite{rademaker2014language}. 

This paper also does not fully consider relevant topics concerning human rights. While we agree with scholars such as \citet{prabhakaran2022human} that human rights provide a useful framework for considering AI ethics, we note that agreement on how to interpret collective and cultural rights is not universal; for some perspectives from human rights scholars see  \cite{jakubowski2016cultural, sanders1991collective, ramcharan1993individual}.


\bibliography{anthology,custom}
\bibliographystyle{acl_natbib}

\newpage
\clearpage
\newpage
\newpage

\section*{Appendix A: Characterizations of the Bible and the Quran}
To select NLP papers for the purposes of examining how they describe the Bible and the Quran, we searched the ACL Anthology for papers mentioning ``the bible is'' or ``the quran is''.  
The Bible is characterized by NLP papers in the ACL Anthology using phrases such as:
\begin{itemize} 
\item `available in electronic form', `most available', `only available', `readily available', `available online' and `often the only resource available for many languages';
\item `parallel', `natural source of parallel data', `ideal source of parallel texts', and `aligned';
\item `short', `large', and `[comparatively] small';
\item `massively multilingual', and `most translated';
\item `central religious text';
\item other descriptions including `narrative of a reasonably straightforward kind', `representative of core vocabulary',
 `one of the most familiar documents'.
 \end{itemize}
 
The Quran is characterized by NLP papers in the ACL Anthology using phrases such as:
\begin{itemize} 
\item  `significant religious text', `[believed by Muslims to be] God's word', `[believed by Muslims to be] God's divine words';
\item `fully diacritized';
\item `written in a unique literary style, close to very poetic language'.        
\end{itemize}

\section*{Appendix B: Selection of ACL Papers}

To select papers to analyze for NLP application domain, we took the first 100 search results on the ACL Anthology for the searches `bible' and `koran'. We omitted 12 `bible' search results from our analysis due to either: 

\begin{itemize}
    \item not being in English (2); 
    \item not being research papers (6, e.g., book reviews, invited talks, or proceedings);
    \item or being duplicates (4)
\end{itemize}
We omitted 36 `koran' search results from our analysis due to either:
\begin{itemize}
    \item not being research papers (16); 
    \item being false positive search results (13, e.g., 10 had typos for `Korean');
    \item only using the word `koran' in references (3);
    \item only using the word `koran' in footnotes (2);
    \item only using the word `koran' in content generated by a model (2).
\end{itemize}

\section*{Appendix C: Application Domains of ACL Papers using the Bible}

\begin{table}[h]
    \centering
    \begin{tabular}{m{6cm}|c}
    \toprule
    Application domain & count \\
    \midrule
         Machine translation&  20\\
         New corpus or resource & 16 \\
         Morpho-syntactic tasks other than POS-tagging and dependency parsing & 9 \\
         POS tagging & 6 \\
         Language modeling/pretraining & 5\\
         Literary analysis of the bible & 4\\
         Bible translation & 3 \\
         Identify text re-use & 2 \\
         Dependency parsing & 2 \\
         Other, including sentiment analysis, named entity, CLIR, patronizing language detection & 21 \\
         \midrule
        Total in selected sample & 88 \\
         \midrule
         Filtered out (see Appendix B) & 12 \\
         \midrule
         Total in sample  & 100 \\
        \bottomrule
    \end{tabular}
    \caption{Application domains of ACL papers using the Bible.}
\end{table}

\newpage
\section*{Appendix D: Application Domains of ACL Papers using the Quran}

\begin{table}[h]
    \centering
    \begin{tabular}{m{6cm}|c}
    \toprule
    Application domain & count \\
    \midrule
         Machine translation&  47\\
         Text generation & 2 \\
         Language models & 2 \\
         NER & 2 \\
         Other, including hate speech and dialogue act classification & 11 \\
         \midrule
        Total in selected sample & 64 \\
         \midrule
         Filtered out (see Appendix B) & 36 \\
         \midrule
         Total in sample & 100 \\
        \bottomrule
    \end{tabular}
    \caption{Application domains of ACL papers mentioning `koran'.}
\end{table}

\section*{Appendix E: A Brief Recent History of Ethics in ACL}

\begin{itemize}
    \item 2017: The First Workshop on Ethics in NLP is held, at EACL \cite{hovy2017proceedings}. 
    \item 2018: The Second Workshop on Ethics in NLP is held, at NAACL-HLT \cite{alfano2018proceedings}.
    \item 2018: ``Ethics and Fairness in NLP'' is a possible submission topic for EMNLP.
    \item 2019: NAACL-HLT has a theme track of data privacy and model bias. 
    \item 2020: ``Ethics and NLP'' becomes a possible submission topic for the ACL conference. This topic, or slight variations of it, are then regular in *ACL and EMNLP conferences from 2021 onwards.
    \item 2020: EMNLP introduces an Ethics Policy, reserving the right to reject papers on ethical grounds. ACL and NAACL follow in 2021, and EACL in 2023.  (There was no EACL in 2022.)
    \item 2021: ACL, NAACL and EMNLP begin allowing extra space in papers for discussions of ethical considerations. Reviewers are asked to consider Ethics Review Questions. EACL follows in 2023.
    \item 2021: EACL has a track on ``Green and Sustainable NLP''. NAACL has a submission topic ``Green NLP''.
    \item 2022: ACL Rolling Review (ARR) emerges as a centralized reviewing service for ACL conferences. It incorporates a Responsible Research Checklist  which includes a section on Ethics and is based on the NeurIPS 2021 paper checklist as well as the work of \citet{rogers2021just} and \citet{dodge2019show}. 
\end{itemize}

%
%
%
%

%
%

\section*{Corrections}
A previous version of this paper stated that the Old Testament is a sacred text of Islam.

\end{document}